\documentclass[review]{elsarticle}
\graphicspath{ {./figures/} }
\usepackage{hyperref}
\usepackage{float}
\usepackage{verbatim} 
\usepackage{apalike}
\restylefloat{figure}
\usepackage{rotating} 
\usepackage{makecell}
\graphicspath{ {./figures/} }
\usepackage[table]{xcolor}
\usepackage{verbatim} 
\usepackage{apalike}
\usepackage{caption}
\usepackage{subcaption}
\usepackage[margin=1in]{geometry} 
\usepackage[]{graphicx}
\usepackage[font=small,skip=0pt]{caption}
\usepackage{enumitem}
\usepackage{booktabs}
\usepackage{longtable}
\usepackage{tabularx}
\usepackage{cancel}
\usepackage{multirow}
\usepackage{supertabular}
\usepackage{algorithmic}
\usepackage{algorithm}
\usepackage{amsmath}
\usepackage{rotating}
\usepackage[normalem]{ulem}
\usepackage[utf8]{inputenc}
\usepackage[T1]{fontenc}
\usepackage{lineno}
\usepackage{array}
\usepackage{makecell}
\usepackage{pdflscape}
\usepackage{afterpage}
\usepackage{capt-of}
\usepackage{lipsum}
\usepackage{changepage}
\usepackage{xcolor}
\usepackage{rotating} 
\usepackage{graphicx}
\newcommand{\blue}[1]{\textcolor{blue}{#1}}
\usepackage{amssymb}
\usepackage{natbib}
\bibpunct{(}{)}{,}{a}{}{;}
\renewcommand{\citep}[1]{(\citealp{#1})}

\biboptions{authoryear}
\begin{document}

\begin{frontmatter}

\begin{titlepage}
\begin{center}
\vspace*{1cm}

\textbf{ \large Exploring Social Media Image Categorization using Large Models with different Adaptation Methods: A Case Study on Cultural Nature’s Contributions to People}

\vspace{1.5cm}

Rohaifa Khaldi$^{a,b}$ (rohaifa@go.ugr.es), Domingo Alcaraz-Segura$^{b,c,d}$ (dalcaraz@ugr.es), Ignacio Sánchez-Herrera$^{f}$ (nachosanchezh@correo.ugr.es), Javier Martínez-López$^{b,d,e}$ (javier.martinez@ugr.es), Carlos Javier Navarro$^{b}$ (carlosnavarro@go.ugr.es), Siham Tabik$^a$ (siham@ugr.es) \\

\hspace{10pt}

\begin{flushleft}
\small  
$^a$ Dept. of Computer Science and Artificial Intelligence, DaSCI, University of Granada, 18071 Granada, Spain \\
$^b$ Interuniversity Institute of Earth System Research in Andalusia (IISTA), University of Granada, Granada, 18071, Spain \\
$^c$ Dept. of Botany, University of Granada, 18071 Granada, Spain \\
$^d$ Andalusian Center for Global Change (ENGLOBA), University of Almería, 04120, Almería, Spain \\
$^e$ Dept. of Ecology, University of Granada, 18071 Granada, Spain \\
$^f$ EDUCA EDTECH Group, Camino de la Torrecilla, 30, 18220, Granada, Spain \\

\vspace{1cm}
\textbf{Corresponding Authors:} \\
\textbf{Rohaifa Khaldi} \\
Dept. of Computer Science and Artificial Intelligence, DaSCI, University of Granada, 18071 Granada, Spain \\
Interuniversity Institute of Earth System Research in Andalusia (IISTA), University of Granada, 18071 Granada, Spain\\
Email: rohaifa@go.ugr.es\\
\textbf{Siham Tabik}\\
Dept. of Computer Science and Artificial Intelligence, DaSCI, University of Granada, 18071 Granada, Spain \\
Email: siham@ugr.es\\

\end{flushleft}        
\end{center}
\end{titlepage}

\title{Exploring Social Media Image Categorization using Large Models {with different Adaptation Methods}: A Case Study on Cultural Nature’s Contributions to People}

\author[label1,label2]{Rohaifa Khaldi \corref{cor1}}
\ead{rohaifa@go.ugr.es}
\author[label2,label3,label4]{Domingo Alcaraz-Segura}
\ead{dalcaraz@ugr.es}
\author[label6]{Ignacio Sánchez-Herrera} 
\ead{nachosanchezh@correo.ugr.es}
\author[label2,label4,label5]{Javier Martínez-López} 
\ead{javier.martinez@ugr.es}
\author[label2]{Carlos Javier Navarro}
\ead{carlosnavarro@go.ugr.es}
\author[label1]{Siham Tabik \corref{cor1}}
\ead{siham@ugr.es}

\cortext[cor1]{Corresponding author.}
\address[label1]{Dept. of Computer Science and Artificial Intelligence, DaSCI, University of Granada, 18071 Granada, Spain}

\address[label2]{Interuniversity Institute of Earth System Research in Andalusia (IISTA), University of Granada, 18071 Granada, Spain}

\address[label3]{Dept. of Botany, University of Granada, 18071 Granada, Spain}

\address[label4]{Andalusian Center for Global Change (ENGLOBA), University of Almería, 04120, Almería, Spain}

\address[label5]{Dept. of Ecology, University of Granada, 18071 Granada, Spain}

\address[label6]{EDUCA EDTECH Group, Camino de la Torrecilla, 30, 18220, Granada, Spain}

%
\begin{abstract}

Social media images provide valuable insights for modeling, mapping, and understanding human interactions with natural and cultural heritage. However, categorizing these images into semantically meaningful groups remains highly complex due to the vast diversity and heterogeneity of their visual content as they contain an open-world human and nature elements. This challenge becomes  greater when categories involve abstract concepts and lack consistent visual patterns.
Related studies involve  human supervision in the categorization process and the lack of public benchmark datasets  make comparisons between these works unfeasible. On the other hand, the continuous advances in large models, including  Large Language Models (LLMs), Large Visual Models (LVMs), and Large Visual Language Models (LVLMs), provide a large space of unexplored  solutions.
In this work 1) we introduce FLIPS a dataset of Flickr images  that capture the interaction between human and nature, and 2) evaluate various solutions based on different types and combinations of large models using various adaptation methods. We assess and report their performance in terms of cost, productivity, scalability, and result quality to address the challenges of social media image categorization.

\end{abstract}


\end{frontmatter}

\section{Introduction} 



An increasing number of studies is intending to use social media data for understanding, mapping and modeling the interaction between humans and different subjects of interests such as invasive species \cite{cardoso2024can}, cultural heritage \cite{qiu2021using,lamas2021monumai}, biodiversity \cite{havinga2023social}, cultural ecosystem services \cite{cardoso2022classifying,moreno2020evaluating} and so forth. 
For example, 
social media platforms such as, Instagram, Flickr, X, Facebook, are widely used by visitors and inhabitants of natural protected areas to share images {\it in situ}. In this context, several recent studies \cite{GATZWEILER2024290, yee2024applying, havinga2023social,lingua2022valuing, moreno2020evaluating} have shown that these georeferenced images provide valuable information over time  for  understanding the interactions between humans and nature, also known as Cultural Ecosystem Services (CES). {Quantifying and mapping the supply of CES is of paramount importance for guiding the decision-making processes of both managers and relevant stakeholders in the realms of biodiversity and ecosystem conservation, environmental monitoring, urban planning, and sustainable land-use management. It is also highly relevant for informing policies that target multiple Sustainable Development Goals (SDGs), by aligning ecological preservation with human well-being, cultural heritage, and equitable access to nature.}  {In addition, Accurate spatial representation of CES not only helps identify areas of high ecological and cultural value but also facilitates the integration of local perceptions and socio-environmental dynamics into policy and planning frameworks. By providing insights into where and how people engage with natural landscapes—through recreation, aesthetic enjoyment, spiritual connection, or heritage appreciation—CES mapping supports more inclusive, evidence-based strategies that balance ecological integrity with societal needs.}

There exist several CES category hierarchy, the one that is evolving towards a standard hierarchy  is named the Common International Classification of Ecosystem Services \href{https://cices.eu/}{CICES revision 5.2}, which distinguishes three coarse categories of CES: ecosystem assets, human-ecosystem co-production, and human assets. However, according to a manual examination of how social media images can much with categories, the authors in
\cite{moreno2020evaluating}  claimed that the legend that aligns more with the visual content of the images include six categories: 1) Cultural-religious, 2) Fauna-flora, 3) Gastronomy, 4)  Landscape-nature, 5) Sport and 6) Urban-Rural.


Grouping  social media images into a number of CES clusters is challenging for multiple reasons:  i) the semantic meaning of several  CES classes is broad, frequently abstract, and may refer to a huge number of different visual contents, ii) the large volume of input images, iii) an important proportion of social media images are irrelevant for CES assessment, for example, images of screenshots of unrelated text or duplicated images among many others. 
Several studies have tackled the task of clustering social media images into a reduced number of CES classes but most approaches either reformulate the problem into a simpler image classification task \cite{havinga2023social,cardoso2022classifying,lingua2022valuing,huai2022using} or include an important manual intervention in the classification process \cite{moreno2020evaluating, yee2024applying}.  

Relevant unsupervised approaches in the field of computer vision are based on two steps, first a neural network is trained to learn inner representations from input images, and then, in a second step, the obtained visual embeddings are clustered using a carefully selected criteria \cite{van2020scan}. However, these methods have been evaluated only on image classification benchmarks, which include less challenging images than the images coming from social media.

On the other hand, the continuous breakthroughs on transformer-based Large Language Models (LLMs), pre-trained on Web-scale text corpora, are significantly extending the capabilities of LLMs to become increasingly general task solvers. 
For example, GPT-3 \cite{brown2020language} achieves good performance not only on all fundamental Natural Language Processing (NLP) tasks but also on tasks that require on-the-fly reasoning  without any sorts of fine-tuning \cite{minaee2024large}. These advances in LLMs have opened the way for the creation of powerful models in computer vision \cite{oquab2023dinov2}. {Pre-trained under the self-supervised learning paradigm, recent Large Vision Models (LVMs) are able to encode a vast amount of visual representations that can be utilized to solve diverse downstream vision tasks. The adaptation process includes the supervised learning (e.g., fine-tuning, lightweight fine-tuning), in-context learning (e.g., few-shot learning, zero-shot learning), or unsupervised learning paradigms (e.g., clustering and dimensionality reduction applied to learned embeddings for class discovery).} Lately, LLMs and LVMs are being used as two fundamental building blocks for designing multi-modal models capable of analyzing both images and language, known as Large Vision-Language Models (LVLMs) \cite{liu2023improvedllava,achiam2023gpt}. Such models allow for a more interactive communication between humans and machines, which make it especially suitable for non AI-expert users.


Addressing the task of social media image categorization using current large models raises several key research questions:

\begin{itemize}
\item {Can large models perform social media image categorization effectively without much customization?}
\item {Which types or combinations of large models (e.g., LVMs, LLMs, and LVLMs) provide the most reliable solutions?}
\item {What adaptation strategies, e.g., fine-tuning, prompting, few-shot learning, are necessary to achieve high-quality categorization performance?}
\end{itemize}

{To explore these questions, we introduce FLIPS, a new dataset for CES categorization, and conduct a comprehensive analysis of both supervised and unsupervised paradigms based on large models—LVMs, LLMs, and LVLMs— to assess their capacity to map social media images into predefined categories.}


The structure of this paper is as follows. Section \ref{S1} provides a review of the relevant literature. In Section \ref{S2}, we give an overview of large models. Section \ref{S3} introduces the CES dataset we created. In Section \ref{S4}, we outline the five approaches employed to address the problem of image categorization. Section \ref{S5} presents and discusses the obtained results. Finally, Section \ref{S6} concludes the paper and highlights directions for future work.


\section{Related work} \label{S1}
{The task of social media image categorization in the context of CES analysis has been approached using various methods, including: (1) manual classification, (2) semi-automatic unsupervised classification, and (3) fully automatic supervised classification. In contrast, other fields have explored fully unsupervised classification approaches to address similar challenges.}

{The first approach, based on manual classification of CES-related images, involves full human intervention. For instance, \cite{moreno2020evaluating} manually analyzed 32,489 Flickr photos from the Sierra Nevada National Park (Granada, Spain) to identify CES categories that align with the visual content. They defined four categories: (1) Landscape and species, (2) Recreation and sports, (3) Culture and heritage, and (4) Others. However, when comparing the assigned categories to responses from an online survey conducted with the photo authors regarding their motivations, a strong agreement was found only for the “Landscape and species” category. This misalignment was attributed to the use of inadequately defined semantic categories.}

{The second approach involves an unsupervised classification method coupled with partial human intervention for the correction and refinement of categories. In \cite{yee2024applying}, the authors simplified CES categories into three broad classes: biotic (interactions between humans and biotic elements of nature), abiotic (interactions with abiotic elements), and human–human (interactions among people). They analyzed 87,090 images using Microsoft’s Azure Computer Vision API to extract embeddings—vectors composed of 5,127 label-confidence pairs. A 50\% confidence threshold was applied to filter out irrelevant labels, followed by hierarchical agglomerative clustering, which produced 430 clusters. These clusters were subsequently refined and manually merged into the three predefined CES classes.}

{The third approach adopts a fully automatic solution based on supervised classification. In this setting, the social media image categorization task is typically reformulated as a standard image classification problem with a reduced number of CES-related categories. Images are then analyzed directly using pretrained Convolutional Neural Networks (CNNs), leveraging their ability to extract high-level visual features. Several studies have followed this strategy to classify images into CES classes, often fine-tuning CNNs on domain-specific data or using them as feature extractors in combination with traditional classifiers \citep{havinga2023social,cardoso2022classifying,lingua2022valuing,huai2022using}.}

{In the computer vision sub-field of fully unsupervised image classification, the proposed approaches aim to automate the entire process of organizing a collection of unlabeled images into meaningful clusters, which can later be assigned semantic interpretations. Existing methods in this context typically involve task-specific models that follow one of two dominant strategies:}

\begin{itemize}
\item {\textbf{Decoupled Feature Learning and Clustering:} In this approach, feature representation learning and clustering are performed in separate stages. During the representation learning phase, a self-supervised learning model is used to generate embeddings.  
This model is typically trained using a pretext task such as predicting image transformations (e.g., rotation, colorization) or using an instance discrimination method (e.g., SimCLR \cite{chen2020simple}). These learned representations are then clustered offline to form image groups that may later be semantically interpreted. One of the most notable methods following this paradigm is SCAN \citep{van2020scan}.}

\item {\textbf{End-to-End Joint Learning:} This strategy involves simultaneous learning of feature representations and clustering assignments in an integrated framework. A CNN is trained alongside a clustering algorithm, with both contributing to the loss function used to update the network. Representative methods include DeepCluster \citep{caron2019unsupervised} and DEC \citep{xie2016unsupervised}.}

\end{itemize}

{While these methods have shown promising results on benchmark datasets of object-centered images, they have not yet been tested on the more complex and diverse task of social media image categorization, particularly in domains such as CES analysis.}

\section{Background} \label{S2}
 
Large models—including LLMs, LVMs, and LVLMs—are characterized not only by their increased scale but also by enhanced capabilities. More notably, they demonstrate emergent abilities—complex behaviors and skills that do not appear in smaller, specialized models \citep{minaee2024large, zhang2024vision}.

\subsection*{Large Language Models: } Natural Language Processing (NLP) has undergone several key breakthroughs that have enabled the development of powerful LLMs. Notable among these advancements are the introduction of the Transformer architecture—particularly the self-attention mechanism—in 2017 \citep{vaswani2017attention}, and the emergence of Self-Supervised Learning (SSL) techniques \citep{oord2018representation}, which allow models to learn rich linguistic representations from vast amounts of unlabeled text data. These innovations have collectively led to LLMs with substantially improved generalization capabilities across a wide range of NLP tasks.
Some of the most impactful LLMs include: 1) BERT (Bidirectional Encoder Representations from Transformers), which was among the first models to achieve strong performance across various NLP benchmarks; 2) RoBERTa (A Robustly Optimized BERT Pretraining Approach) \citep{liu1907roberta}, which builds on BERT with enhanced training strategies and larger datasets; and GPT-3 (Generative Pretrained Transformer 3) \citep{brown2020language}, one of the largest LLMs, with 175 billion parameters, demonstrating remarkable performance across diverse tasks and domains.

LLMs can be adapted to new tasks through various techniques:
1) Full fine-tuning or lightweight fine-tuning (i.e., linear probing), which involves training the model on labeled datasets for specific applications. For example, ChatGPT is based on a fine-tuned version of GPT-4.
2) Instruction tuning, where the model is trained on a diverse set of NLP tasks framed as natural language instructions, as implemented in models like FLAN-T5 and FLAN-PaLM \citep{chung2024scaling}.
3) Prompting, which leverages few-shot learning or prompt engineering to guide the model’s responses without modifying its parameters.
4) Retrieval-Augmented Generation (RAG) \citep{lewis2020retrieval}, which enhances model outputs by incorporating relevant information retrieved from an external knowledge base.
{5) Compression-Augmented Generation (CAG), a lightweight alternative to RAG, which uses compressed representations of external knowledge to guide the generation process.} To reduce the computational cost of adapting large models, model distillation techniques are often employed, resulting in compact yet effective variants such as DistilBERT \citep{sanh2019distilbert}.

\subsection*{Large Vision Models: }
{Following the success of LLMs, LVMs have emerged as powerful foundational models pretrained on massive image datasets using SSL techniques. These models learn to generate rich, general-purpose visual representations that can be effectively adapted to a variety of downstream vision tasks via fine-tuning, few-shot, or even zero-shot learning approaches.}

{Among the most influential LVMs are CLIP \citep{radford2021learning}, which aligns visual and textual representations to enable robust zero-shot classification, and DINOv2 \citep{oquab2023dinov2}, which is widely recognized as a state-of-the-art self-supervised vision model. DINOv2 was trained on large, curated, and diverse image datasets, allowing it to produce high-quality, transferable features suitable for a wide array of applications, including image classification, segmentation, and retrieval.}

\subsection*{Large Vision Language Models: }
{LVLMs have emerged as a natural evolution of integrating LLMs with LVMs, enabling seamless multimodal interaction. By utilizing machine-generated instruction-following data during training, these models facilitate more dynamic and interactive communication between humans and machines. LVLMs are capable of processing both visual and textual inputs, generating coherent and contextually relevant responses in natural language.}

{These models serve as  general-purpose problem solvers, demonstrating good performance across various tasks—particularly when prompt engineering is used to guide the model, and when the task aligns with the domain on which the model was trained.}{Notable LVLMs include BLIP \citep{li2022blip}, one of the earliest unified vision-language encoder-decoder architectures; LLaVA (Large Language and Vision Assistant) \citep{liu2023improvedllava}, an end-to-end multimodal model that connects a vision encoder to a language model for versatile visual-language understanding; and GPT-4 \citep{achiam2023gpt}, a proprietary foundation model capable of multimodal reasoning, accessible exclusively via APIs such as ChatGPT.}







\section{Data collection} \label{S3}

We created a new dataset named \href{https://zenodo.org/records/14035313}{FLIPS} (Flickr Images from Mountain National Parks) containing photographs of human interactions with nature \citep{khaldi_2024_14035313}. This dataset was compiled from pictures shared between 2015 and 2022 through Flickr social-media platform, from eight mountain national parks: Seven parks across Spain (Teide, Sierra Nevada, Sierra de las Nieves, Guadarrama, Picos de Europa, Ordesa y Monte Perdido, and Aigüestortes i Estany de Sant Maurici) and one in Portugal (Peneda-Gerês). 

\begin{figure}[H]
  \begin{center}
    \includegraphics[width=15cm]{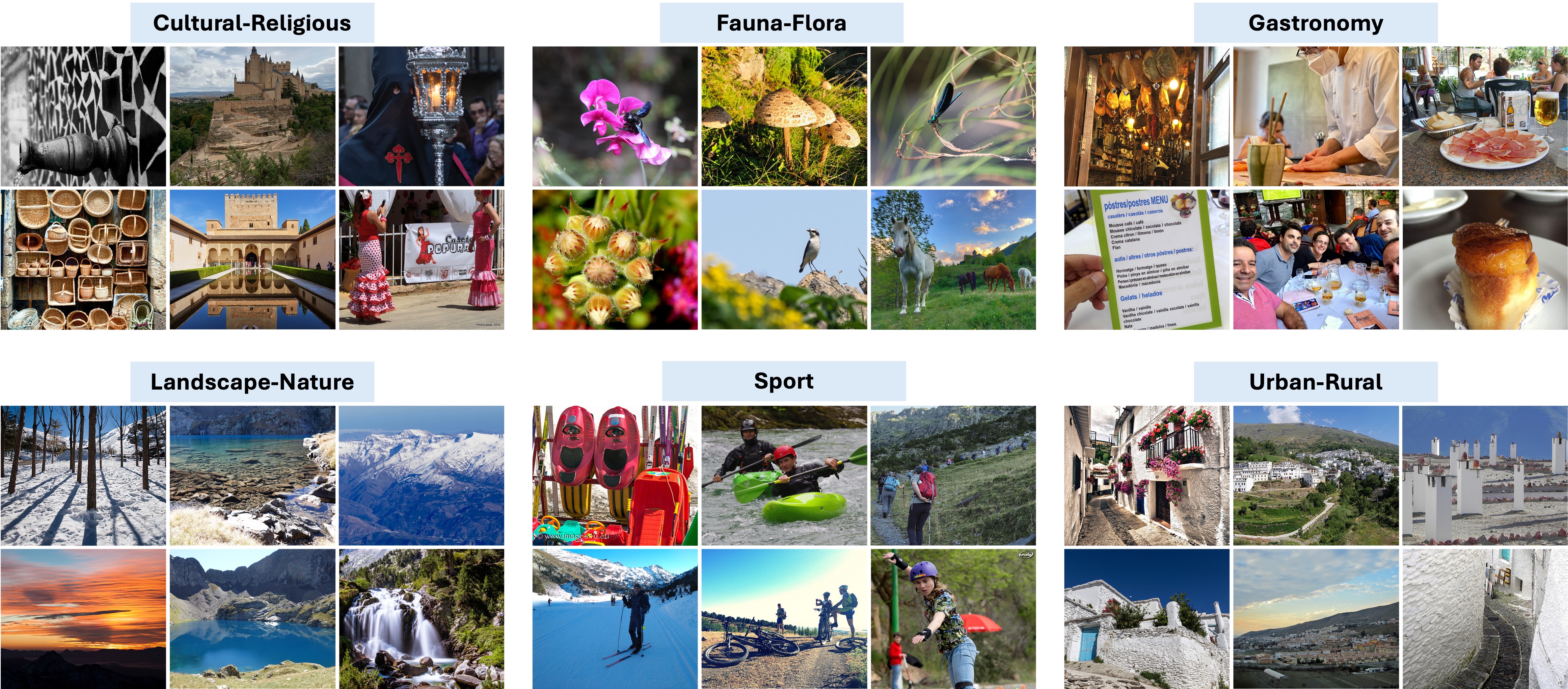} 
  \end{center}
  \caption{Six examples of  each CES category.}
  \label{fig1}
\end{figure}


To ensure high data quality, we implemented a rigorous filtering process to remove  noisy images, for example images containing only text or screen shots. Following the methodology of \cite{moreno2020evaluating}, we carefully selected a representative number of dissimilar pictures for each class. The obtained FLIPS dataset contains $960$ images distributed evenly across six CES classes: 1) Cultural-Religious, 2) Fauna-Flora, 3) Gastronomy, 4) Landscape-Nature, 5) Sport, and 6) Urban-Rural.
{These six categories were established based on insights gained from multiple exploratory experiments. While they differ from those used in \cite{moreno2020evaluating} original classification, our selected classes are more representative thereby enabling more reliable learning and improved generalization by vision models.}

\section{Study design} \label{S4}
\begin{figure}[H]
  \begin{center}
    \includegraphics[width=14cm]{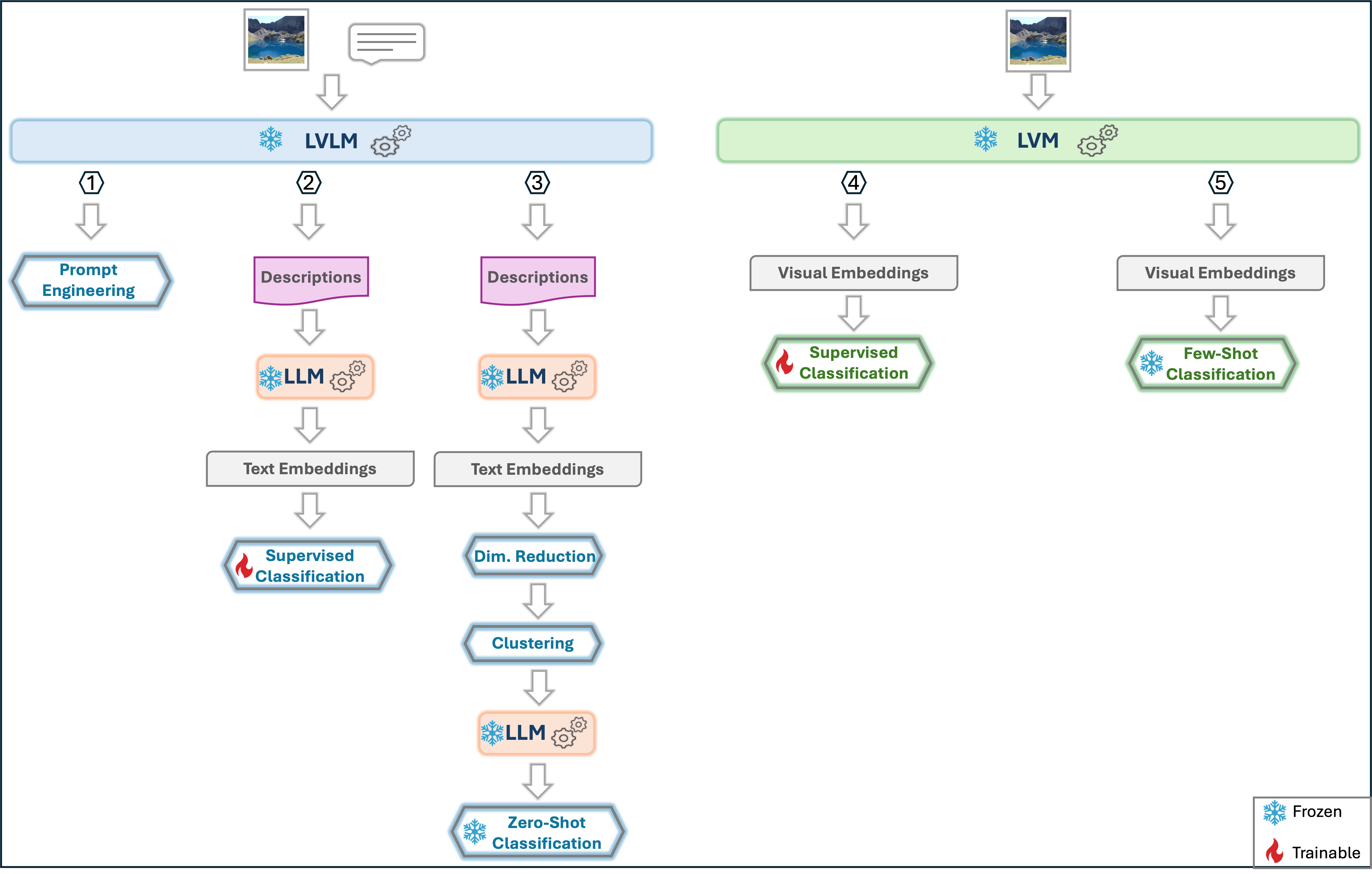} 
  \end{center}
  \caption{{High-level illustration of the five proposed approaches, indicated as (1), (2), (3), (4) and (5), to address social media images categorization using large models. See Table \ref{tab1} for more information about the type of the used  models and adaptation methods.}}
  \label{fig2}
\end{figure}

To assess the capability of large models in categorizing social media images into CES classes—and to determine the most effective model type, combination, and adaptation strategy for achieving a high-quality solution—we conduct a comparative evaluation of five approaches. These approaches are based on one or multiple state-of-the-art LLMs, LVMs, and LVLMs, and are examined across three learning paradigms: supervised learning, in-context learning, and a hybrid of unsupervised and in-context learning. 
The specific adaptation methods employed within each paradigm are detailed below (Figure \ref{fig2}):

\begin{itemize}
\item \textbf{Supervised learning adaptation methods:} {We assess the effectiveness of lightweight fine-tuning, specifically through linear probing, as a widely used adaptation method.}

\item \textbf{In-context learning adaptation methods:} {We evaluate two different adaptation methods, namely prompting through extensive prompt engineering, and few-shot learning with example-based inputs.} 

\item \textbf{Hybrid unsupervised and in-context learning adaptation method:} {We explore a combined approach involving dimensionality reduction followed by clustering of model embeddings, paired with zero-shot learning for classification.}
\end{itemize}

In this context, we proposed five approaches, labeled as Approach (1), (2), (3), (4) and (5), a graphical illustration of these approaches  highlighting the used learning paradigm, adaptation method, and large model types,  is provided  in Figure \ref{fig2}. A more detailed description of each approach is given in Table \ref{tab1}. Hereafter, we describe all the proposed approaches evaluated under each adaptation method.


\begin{table}[h]
\centering
\caption{{Summary of large models and adaptation methods used to develop the solutions for social media images categorization. Larger size (i.e., larger number of parameters) indicates higher computational requirements. B stands for billions and M for millions.}}

\label{tab1}
\resizebox{\textwidth}{!}{%
\begin{tabular}{lcclcc}
\hline
\multicolumn{1}{l}{\textbf{\thead{Learning paradigm}}} & \multicolumn{1}{c}{\textbf{\thead{Adaptation method}}} & 
\multicolumn{1}{c}{\textbf{\thead{Approach}}} & 
\multicolumn{1}{c}{\textbf{\thead{Model}}} & {\textbf{\thead{Size}}} & {\textbf{\thead{Proprietary}}}\\ \hline

\multirow{9}{*}{\thead{\bf Supervised}} & \multirow{9}{*}{\thead{Lightweight \\Fine-Tuning}} & \multirow{6}{*}{\thead{2}} & \thead{LLaVA-1.5 + BERT}& \thead{7B + 110M} & \thead{No}\\
& & & \thead{LLaVA-1.5 + DistilBERT} & \thead{7B + 66M} & \thead{No}\\
& & & \thead{LLaVA-1.5 + RoBERTa} & \thead{7B + 125M} & \thead{No}\\
& & & \thead{BLIP + BERT} & \thead{400M + 110M} & \thead{No}\\
& & & \thead{BLIP + DistilBERT} & \thead{400M + 66M} & \thead{No}\\
& & & \thead{BLIP + RoBERTa} & \thead{400M + 125M} & \thead{No}\\ \cline{4-6}
&  & \multirow{3}{*}{\thead{4}} & \thead{DINOv2 - ViT-S/14} & \thead{21M}  & \thead{No}\\ 
&  & & \thead{DINOv2 - ViT-B/14} & \thead{86M}  & \thead{No}\\ 
&  & & \thead{DINOv2 - ViT-L/14} & \thead{0.3B}  & \thead{No}\\ 
\hline

\multirow{6}{*}{\thead{\bf In-Context}} & \multirow{3}{*}{\thead{Zero-Shot with Prompting}} & \multirow{3}{*}{\thead{1}} & \thead{LLaVA-1.5} & \thead{7B} & \thead{No} \\
& & &  \thead{GPT-4o} & \thead{175B} & \thead{Yes}\\
& & &  \thead{GPT-4o-mini} & \thead{1.5B} & \thead{Yes}\\ 
\cline{4-6}
& \multirow{3}{*}{\thead{Few-Shot}} & \multirow{3}{*}{\thead{5}} & \thead{DINOv2 - ViT-S/14} & \thead{21M}  & \thead{No}\\ 
&  & & \thead{DINOv2 - ViT-B/14} & \thead{86M}  & \thead{No}\\ 
&  & & \thead{DINOv2 - ViT-L/14} & \thead{0.3B}  & \thead{No}\\ 
\hline

\multirow{1}{*}{\thead{\textbf{Unsupervised + In-Context}}} & \thead{Dimension Reduction \\+ Clustering \\+ Zero-Shot\\} & \multirow{1}{*}{\thead{3}} & \thead{LLaVA-1.5 + SBERT \\+ Flan-T5} & \thead{7B + 220M + 110M} & \thead{No}\\ 
\hline

\end{tabular}
}
\end{table}

\subsection{Supervised learning adaptation methods:}
{We assess under this paradigm the effectiveness of two large models-based approaches,  Approach  (2) and Approach (4) (see Figure \ref{fig2}) using lightweight fine-tuning adaptation which consists of training the last fully connected layer of the model.}
\begin{itemize}
    \item Approach (4)  employs a pretrained DINOv2 LVM model \citep{oquab2023dinov2} combined with a supervised SoftMax classifier. The method ingests an input image and uses the pretrained LVM to generate visual embeddings. These embeddings are then processed by the classifier to predict the class with the highest probability. Only the classifier head is trained, while the backbone remains frozen.

    \item {Approach (2) utilizes a conjunction of pretrained LVLM and LLM. The solution unfolds as follows: 
i)  We first reformulate the problem as a scene captioning task in which the LVLM model is instructed  to generate a brief description ("Describe the image. Keep your response short."). We specifically request the model to provide a short response to make it  focus on the relevant aspects of the image according to its judgment. We evaluate two popular architectures: LLaVA-1.5 \citep{liu2024improved} and BLIP \citep{li2022blip}. }
\end{itemize}

\begin{figure}[H]
  \begin{center}
    \includegraphics[width=11cm]{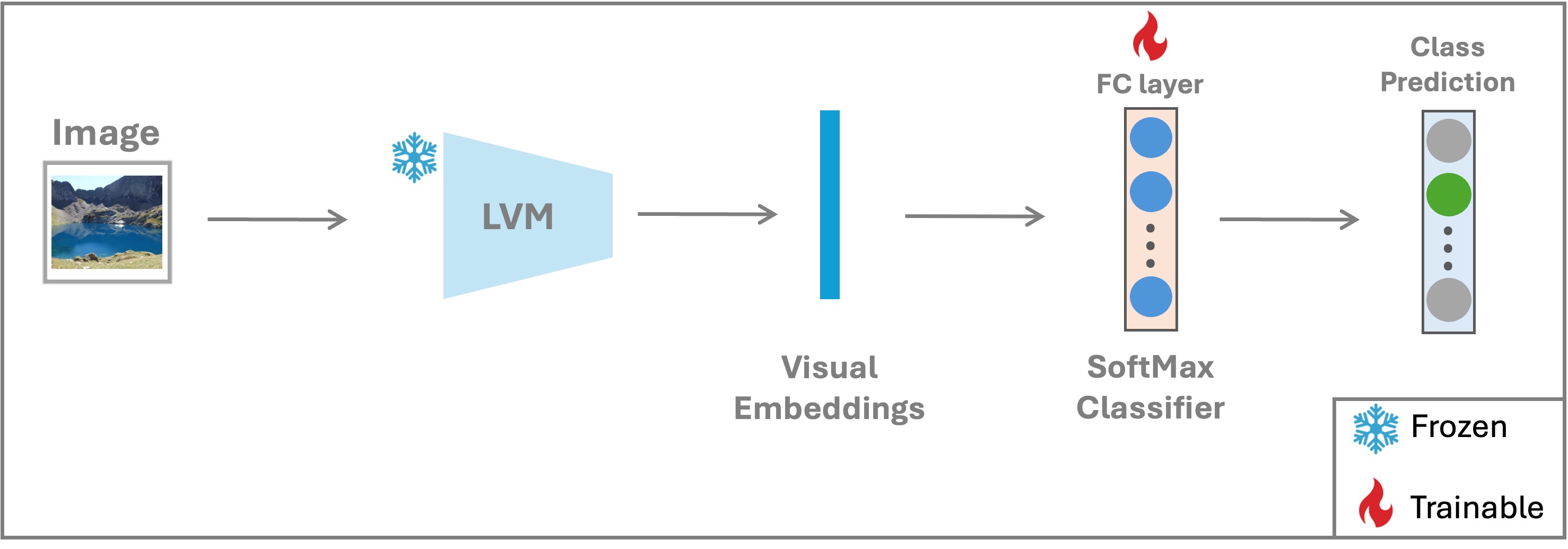} 
  \end{center}
  \caption{{{ Illustration of Approach (4)}: Supervised learning-based solution involving LVM adapted using lightweight fine-tuning method. Frozen implies that the model was directly applied without any training.}}
  \label{ex4}
\end{figure}

\begin{figure}[H]
  \begin{center}
    \includegraphics[width=12cm]{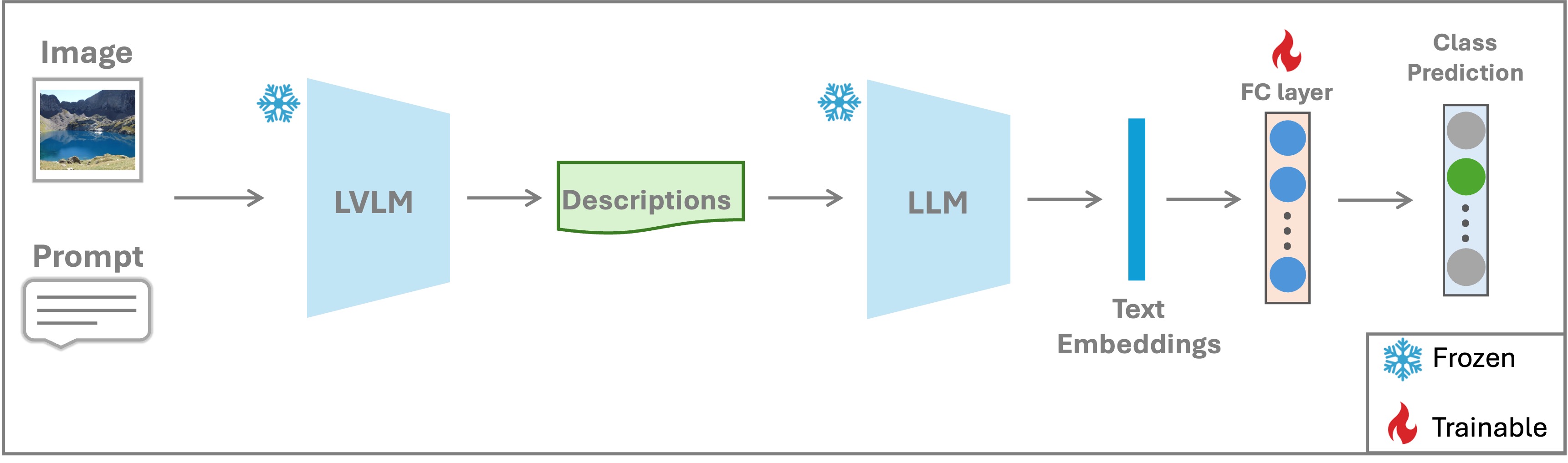} 
  \end{center}
  \caption{{{Illustration of Approach (2): }Supervised learning-based solution involving LVLM combined with LLM adapted using lightweight fine-tuning method. Frozen implies that the model was directly applied without any training.}}
  \label{ex2}
\end{figure}

ii) We then use the LLM to generate text embeddings for these descriptions, experimenting with three different architectures: BERT \citep{devlin2018bert}, DistilBERT \citep{sanh2019distilbert}, and RoBERTa \citep{liu2019jingfei}. iii) Finally, we train the head-layer to classify the descriptions into our predefined CES categories.

\subsection{In-context learning adaptation methods:}

\begin{figure}[H]
  \begin{center}
    \includegraphics[width=12cm]{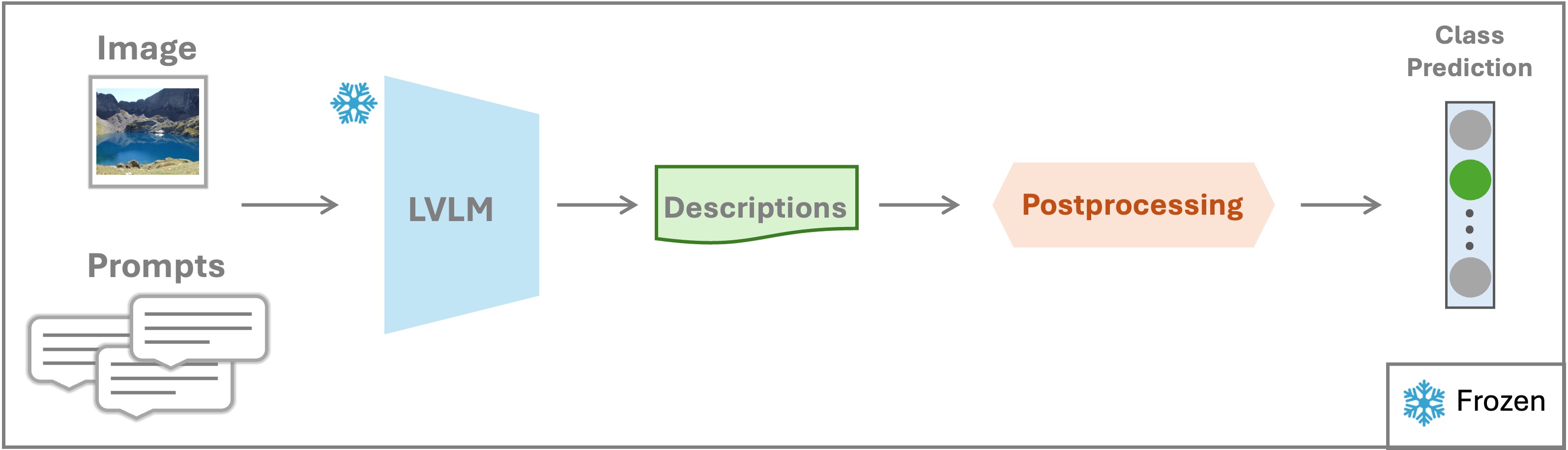} 
  \end{center}
  \caption{Illustration of Approach (3): In-context learning-based solution using LVLM adapted with prompting method. Frozen implies that the model was directly applied without any training.}
  \label{ex1}
\end{figure}

\begin{figure}[H]
  \begin{center}
    \includegraphics[width=9cm]{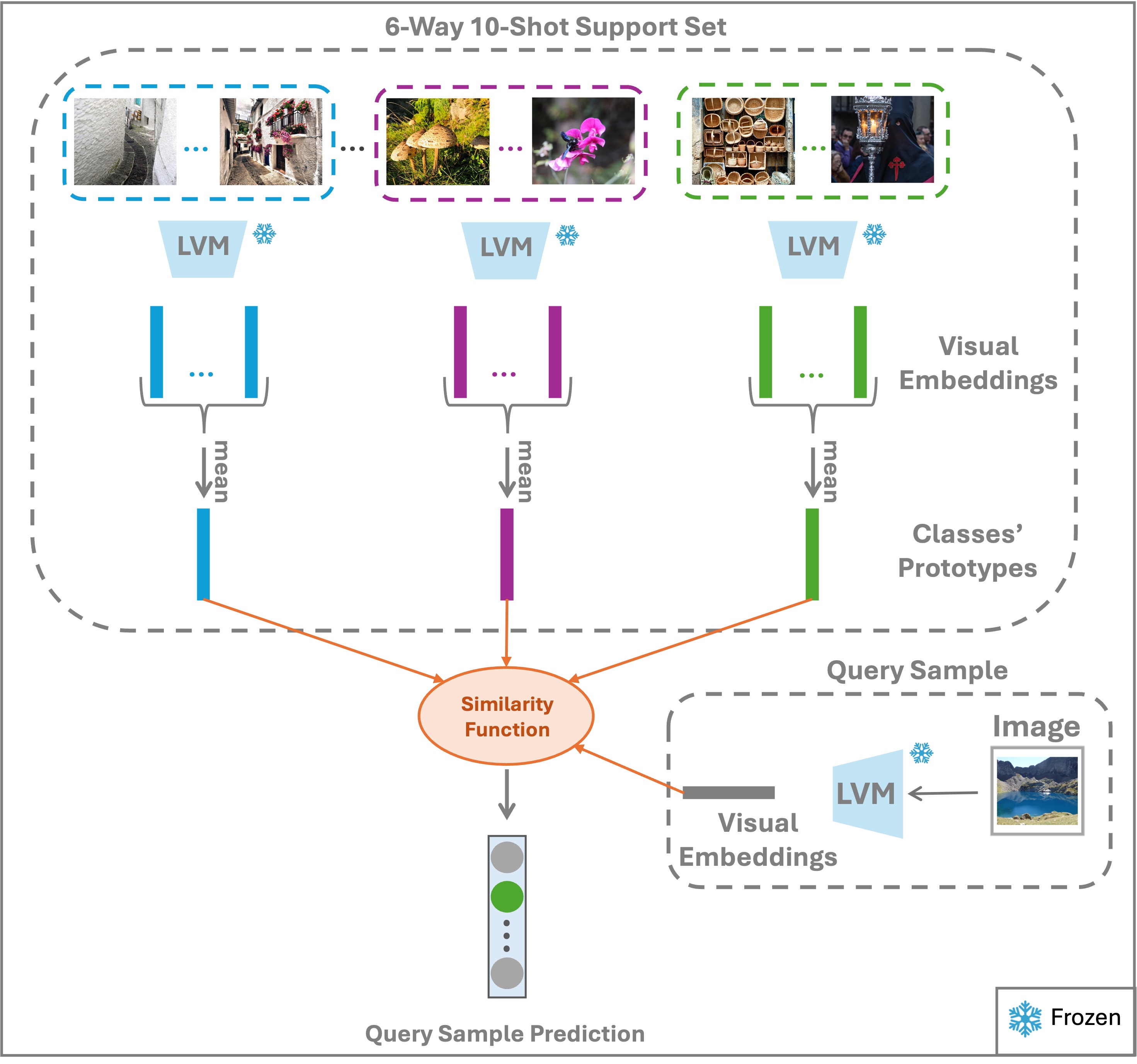} 
  \end{center}
  \caption{{Illustration of Approach (5): In-context learning-based solution using LVM adapted with few-shot learning method. Frozen implies that the model was directly applied without any training.}}
  \label{ex5}
\end{figure}

{We evaluate two approaches (1) and (5) (see Figure \ref{fig2}) using two large models, i.e., LVM and LVLM,  adapted to our downstream task through two different methods of in-context learning, namely prompting using predefined templates, and few-shot learning.}
\begin{itemize}
    \item Approach (1) addresses the downstream task of image categorization as a Visual Question Answering (VQA) task in which a pretrained LVLM is requested to  assign each input image with one category of a list of categories (Figure \ref{ex1}). An extensive analysis of the prompt was carried out to find the most suitable one. 


{A post-processing step is applied to the model’s output to extract the CES class from the generated text.  Two LVLM models are tested (Table \ref{tab1}):  the public LLaVa-1.5 and two variants of the proprietary  GPT-4\footnote{The black-box GPT-4 \citep{achiam2023gpt} is used through ChatGPT API}: GPT-4o which is an improved version of GPT-4 and GPT-4o-mini which is a cost-efficient smaller version of GPT-4o, under two different modes (i.e., batch and no batch). In the batch mode, multiple inputs are sent to the model in a single API call, while the no-batch mode processes one input image at a time. These models are applied directly to the input images, without any additional training, using various simple and extended prompts. The two prompts (prompt 1 and 2) that provide the best performances are shown in Table \ref{prompts}.}

\item {In Approach (5),  we utilize a pretrained LVM using few-shot classification (Figure \ref{ex5}). We use two data partitions: i) a support set with different configurations of shots (i.e., number of samples), ranging from 1 to 10, and ii) a query set consisting of all the remaining samples from the original data that do not belong to the support set.
To assess the model's stability concerning the support set samples, we generate 30 random support sets for each shot configuration and compute the average model performance. 
The method operates as follows: i) we create a support set containing six classes with a specified number (i.e., ranging from 1 to 10) of shots per class; ii) we apply the LVM to generate visual embeddings for both the support and query set samples; iii) we compute a prototype for each class in the support set by averaging the embeddings of the samples in that class; iv) we compare the embedding vector of the query sample with the prototypes' embeddings using a cosine similarity function normalized by the Softmax function; finally, v) we assign to the query sample the class label of the closest prototype (i.e., the one with the highest Softmax value). For this approach, we employ the LVM DINOv2.}

\end{itemize}

\subsection{Hybrid unsupervised and in-context learning adaptation method:}

{In Approach (3) (see Figure \ref{fig2}), we leverage a combination of LVLM and LLMs adapted through a hybrid method combining unsupervised learning (i.e., dimensionality reduction followed by clustering) and in-context learning (i.e., zero-shot learning).}

\begin{figure}[H]
\centering
  \begin{center}
    \includegraphics[width=12cm]{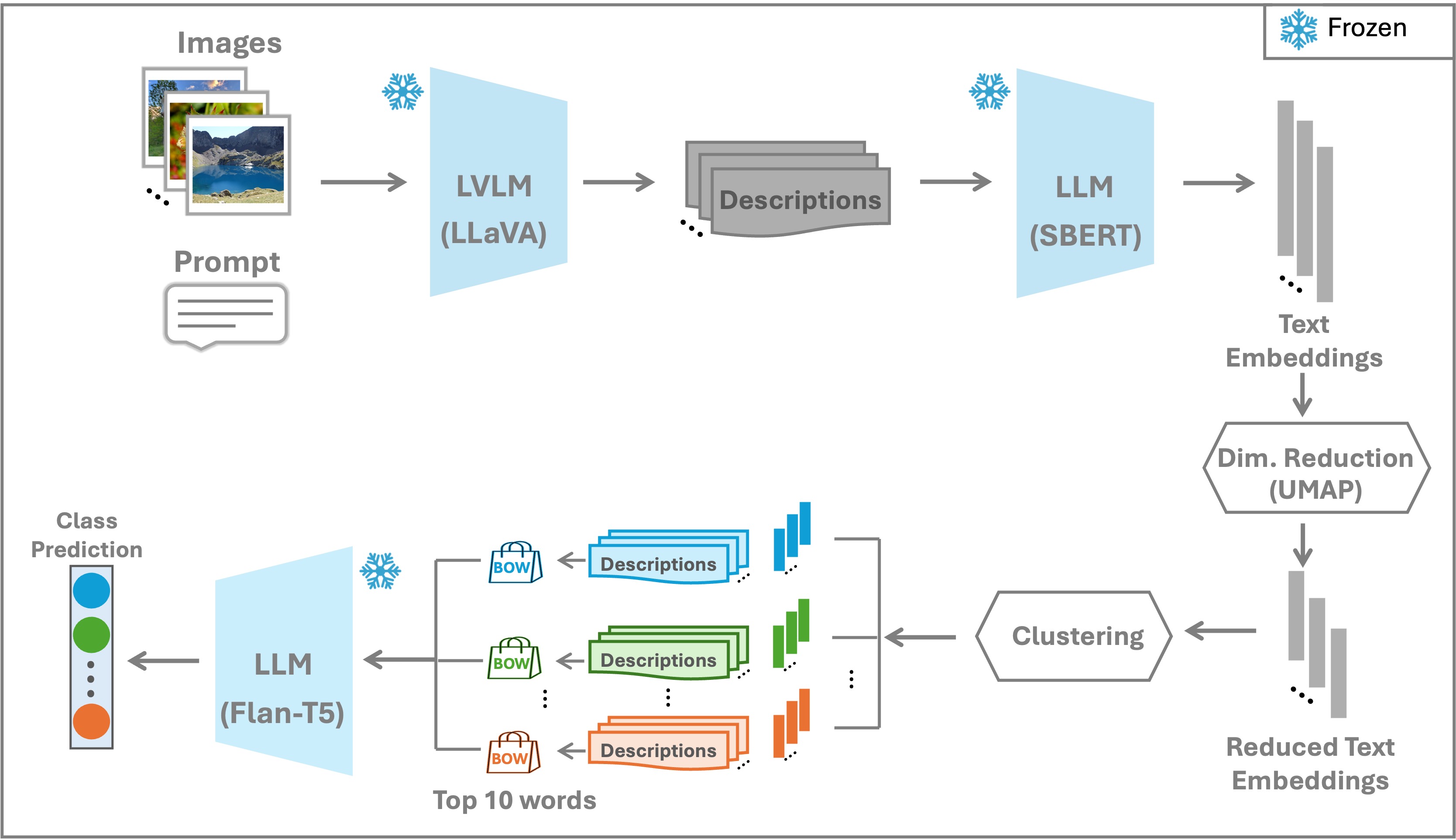} 
  \end{center}
  \caption{{Illustration of Approach (3): Solution combining LVLM and LLMs adapted through a hybrid unsupervised and in-context learning. Frozen implies that the model was directly applied without any training. }}
  \label{ex3}
\end{figure}

{The solution is described as follows (see Figure \ref{ex3}): i) We use the LVLM LLaVA-1.5 to generate descriptions for each image. { We use the same prompt as in Approach (2)} ii) These descriptions are then converted into text embeddings using the LLM Sentence-BERT (SBERT) \citep{reimers2019sentence}. iii) To mitigate the curse of high dimensionality, we apply UMAP \citep{mcinnes2018umap}, a state-of-the-art dimensionality reduction model. This model projects these embeddings into a lower-dimensional space while preserving both local and global data structure. iv) Clustering is performed on the reduced embeddings using two different clustering models, KMeans and HDBSCAN \citep{mcinnes2017hdbscan}. v) For each cluster, we have a collection of image descriptions. Then, we tokenize these descriptions and subsequently identify the ten most representative words based on Term Frequency (TF) and Inverse Document Frequency (IDF) scores. These top ten words constitute the Bag of Words (BOW) for each cluster. vi) Finally, we use the LLM Flan-T5 (google/flan-t5-large) \citep{chung2024scaling} adapted through zero-shot learning to map each cluster to our predefined classes based on the generated BOW per cluster and a specific prompt structure. Flan-T5, having been fine-tuned with instruction-based learning across various tasks, is well-suited for zero-shot text classification \citep{mann2020language}.}

\subsection{Experimental setup}
The choice and configuration of hyperparameters varied across approaches, depending on both the number of models involved and the type of learning paradigm adopted (i.e., with or without training). For models adapted through supervised learning, LLMs are fine-tuned for 5 epochs with a batch size of 16, a learning rate of $2\times10^{-5}$, and a weight decay of 0.01. LVMs, on the other hand, are fine-tuned for 100 epochs with a learning rate of $2\times10^{-3}$.
In the in-context learning setup, LVLMs are evaluated under various batch size configurations—batch size of 1 for LLaVA-1.5, and batch sizes of 1 and 50 for GPT-4 variants. LVMs adapted via few-shot learning use a cosine similarity function, and experiments are conducted with different numbers of shots, ranging from 1 to 10.
For models adapted through unsupervised and in-context learning, UMAP is used for dimensionality reduction, optimized with 15 neighbors, 20 output dimensions, and the cosine distance metric. For clustering, HDBSCAN achieves the best results using a minimum cluster size of 20, Euclidean distance, and the leaf cluster selection method, while KMeans is configured to generate 6 clusters.

An extensive prompt engineering process is conducted for LVLM-based methods, resulting in two optimal prompt formats (Table \ref{prompts}): a simple prompt (Prompt 1) and an extended prompt (Prompt 2).

\begin{table}[H]
\caption{The best two prompts used in LVLM-based approaches: (1) a simple prompt and (2) an extended prompt that included a definition for each class generated by ChatGPT-4.}
\centering
\label{prompts}
\resizebox{\textwidth}{!}{%
\begin{tabular}{cl}  
\hline
\textbf{Prompt id} & \textbf{Prompt description} \\ \hline
1 & \makecell[l]{Classify the image into one of these categories: Cultural-Religious, Fauna-Flora, Gastronomy, \\ Landscape-Nature,
Sports, or Urban-Rural.}\\ \hline
2 & \makecell[l]{Classify the image into one of these categories: Cultural-Religious, Fauna-Flora, Gastronomy, \\ Landscape-Nature,
Sports, or Urban-Rural. \\The definitions for each category are as follows:\\
\textbf{Cultural-Religious:} The image depicts religious symbols, cultural artifacts, traditions, ceremonies, \\  or anything related to culture and belief systems.\\
\textbf{Fauna-Flora:} The image features animals (fauna) or plants (flora) in any environment.\\
\textbf{Gastronomy:} The image is related to food, cooking, culinary experiences, or dining.\\
\textbf{Landscape-Nature:} The image contains natural landscapes, such as mountains, rivers, forests, \\ or other  untouched environments.\\
\textbf{Sports:} The image shows physical activities, competitions, or sports equipment related to athletic\\ endeavors.\\
\textbf{Urban-Rural:} The image captures cityscapes, villages, rural settings, buildings, or any human-made \\  environments.} \\ \hline
\end{tabular}
}
\end{table}

\section{Experimental results and discussion } \label{S5}

{In this section, we first evaluate the performance of various large model-based solutions within each adaptation method independently, then we perform a comparative analysis of the best models across the different adaptation strategies.}

\begin{table}[h]
\centering
\caption{{Test results of various large model-based solutions adapted to the task of social media image categorization using the lightweight fine-tuning method. The best-performing model is highlighted in bold. Model sizes are denoted as S (small), B (base), and L (large).}}
\label{a2}
\begin{tabular}{clrrr} \hline
\multicolumn{1}{c}{\textbf{Adaptation method}} & \multicolumn{1}{c}{\textbf{Model}} & \multicolumn{1}{c}{\textbf{Precision}} & \multicolumn{1}{c}{\textbf{Recall}} & \multicolumn{1}{c}{\textbf{Accuracy}} \\ \hline
\multirow{9}{*}{Lightweight Fine-Tuning} & LLaVA-1.5 + BERT  & 94.95 & 94.79 & 94.79 \\
& LLaVA-1.5 + DistilBERT  & 95.08 & 95.00 & 95.00 \\
& LLaVA-1.5 + RoBERTa  & 94.92 & 94.79 & 94.79 \\
& BLIP + BERT & 88.73 & 88.65 & 88.65 \\
& BLIP + DistilBERT & 88.24 & 88.02 & 88.02 \\
& BLIP + RoBERTa & 88.49 & 88.23 & 88.23 \\ 

& DINOv2 - ViT-S/14 & 94.77 & 94.58 & 94.58 \\
& DINOv2 - ViT-B/14 & 96.85 & 96.77 & 96.77 \\
& \bf DINOv2 - ViT-L/14 & \bf 97.20 & \bf 97.08 & \bf 97.08 \\ 
\hline
\end{tabular}
\end{table}
{Under the supervised learning paradigm, we assessed a range of large models, including standalone LVMs as well as combinations of LVLMs and LLMs, all adapted to the downstream task using lightweight fine-tuning adaptation method. The evaluation results are summarized in Table \ref{a2}.
The highest classification accuracy was achieved by the LVM DINOv2, utilizing the ViT-L/14 (i.e., large Vision Transformer) backbone, with an accuracy of 97.08\%. This performance surpasses all tested combinations of LVLMs and LLMs. Among those, the best-performing pair—LLaVA-1.5 coupled with DistilBERT—achieved an accuracy of 95.00\%.}

\begin{table}[H]
\centering
\caption{{Test results of various large model-based solutions for social media image categorization using in-context learning adaptation methods. The evaluated methods include prompting with Prompt 1 (simple format) and Prompt 2 (extended format), as well as 10-shot few-shot learning. The best-performing model is shown in bold. Model sizes are denoted as S (small), B (base), and L (large).}}
\label{a1}
\begin{tabular}{clclrr} \hline

\multicolumn{1}{c}{\textbf{Adaptation method}} & \multicolumn{1}{c}{\textbf{Model}} & \multicolumn{1}{c}{\textbf{Prompt}} & \multicolumn{1}{c}{\textbf{Precision}} & \multicolumn{1}{c}{\textbf{Recall}} & \multicolumn{1}{c}{\textbf{Accuracy}} \\ \hline

\multirow{10}{*}{Zero-Shot Prompting} & \multirow{2}{*}{LLaVA-1.5} & 1 & 87.02 & 84.48 & 84.48 \\
& & 2 & 86.05 & 80.10 & 80.10 \\ \cline{2-6}

 & \multirow{2}{*}{\bf GPT-4o (no batch)} & \bf 1 & \bf 91.04 & \bf 88.85 & \bf 88.85 \\
& & 2 & 89.90 & 86.98 & 86.98 \\ \cline{2-6}

 & \multirow{2}{*}{GPT-4o (batch)} & 1 & 84.82 & 80.52 & 80.52 \\
& & 2 & 90.19 & 87.50 & 87.50 \\ \cline{2-6}

 & \multirow{2}{*}{GPT-4o-mini (no batch)} & 1 & 90.41 & 87.92  & 87.92 \\
& & 2 & 90.15 & 88.02 & 88.02 \\ \cline{2-6}

 &  \multirow{2}{*}{GPT-4o-mini (batch)} & 1 & 90.15 & 87.71 & 87.71 \\
& & 2 & 90.24 & 88.12 & 88.12 \\ 

\hline

\multirow{3}{*}{Few-Shot} & \bf DINOv2 - ViT-S/14 & & \textbf{83.56} & \textbf{83.54} & \textbf{83.99} \\
& DINOv2 - ViT-B/14 & & 80.61 & 80.40 & 80.67 \\
& DINOv2 - ViT-L/14 & & 79.89 & 79.62 & 79.96 \\ \hline

\end{tabular}
\end{table}


{Under the in-context learning adaptation methods, we evaluated a range of large models using two distinct strategies: prompting for LVLMs and few-shot learning for LVMs. The results are summarized in Table \ref{a1}.
Using the prompting approach, the best performance was achieved by the LVLM GPT-4o, operating in no-batch mode with the simple prompt format (Prompt 1). In contrast, for the few-shot learning method, the highest accuracy was obtained by the LVM DINOv2 with the ViT-S/14 (small Vision Transformer) backbone, reaching 83.99\% accuracy using 10-shot learning.
These findings indicate that, for the task of categorizing social media images into CES classes under in-context learning settings, the most effective solution was the large model GPT-4o, adapted via prompt engineering.}


\begin{table}[h]
\centering
\caption{{Test results of a large model-based solution for social media image categorization using hybrid unsupervised and in-context learning adaptation method. The best-performing model is shown in bold.}}
\label{a3}
\resizebox{\textwidth}{!}{%
\begin{tabular}{cllrrr} \hline

\multicolumn{1}{c}{\textbf{Adaptation method}} & \multicolumn{1}{c}{\textbf{Model}} & \multicolumn{1}{c}{\textbf{Clustering model}} & \multicolumn{1}{c}{\textbf{Precision}} & \multicolumn{1}{c}{\textbf{Recall}} & \multicolumn{1}{c}{\textbf{Accuracy}} \\ \hline

\multirow{2}{*}{\thead{Dimension Reduction + \\Clustering + Zero-Shot}} & \multirow{2}{*}{\thead{LLaVA-1.5 + SBERT\\ + Flan-T5}} & \multicolumn{1}{c}{\textbf{KMeans}} & 66.08 & \textbf{72.60} & \textbf{72.60} \\ 
&  & \multicolumn{1}{c}{HDBSCAN} & \textbf{85.45} & 70.73 & 70.73 \\
\hline
\end{tabular}
}
\end{table}

{In Table \ref{a3}, we present the results of the hybrid unsupervised and in-context learning adaptation method, which combines dimensionality reduction and clustering of model embeddings, followed by zero-shot classification. This approach integrates the LVLM LLaVA-1.5 with the LLMs SBERT and Flan-T5, evaluated using two clustering algorithms: KMeans and HDBSCAN.
The highest classification accuracy was achieved with KMeans, reaching 72.60\%, while HDBSCAN yielded a slightly lower accuracy of 70.73\%. Interestingly, HDBSCAN outperformed KMeans in terms of precision, achieving 85.45\% compared to 66.08\%, suggesting that HDBSCAN, while less accurate overall, produced more confident and precise predictions.}

\begin{table}[h]
\centering
\caption{{Test results of the best-performing large model-based solution for each adaptation method. The best-performing model is shown in bold.}}
\label{tab3}
\begin{tabular}{llrrr} \hline
\multicolumn{1}{c}{\textbf{Adaptation method}} & \multicolumn{1}{c}{\textbf{Model}} & \multicolumn{1}{c}{\textbf{Precision}} & \multicolumn{1}{c}{\textbf{Recall}} & \multicolumn{1}{c}{\textbf{Accuracy}} \\ \hline
\thead{\bf Lightweight Fine-Tuning} & \thead{\bf DINOv2 - ViT-L/14} & \bf 97.20 & \bf97.08 & \bf 97.08 \\ 
\thead{Zero-Shot Prompting} & \thead{GPT-4o (No Batch)} & 91.04 & 88.85 & 88.85\\
\thead{Few-Shot} & \thead{DINOv2 - ViT-S/14} & 83.56 & 83.54 & 83.99 \\
\thead{Dimension Reduction + \\Clustering + Zero-Shot} & \thead{KMeans + LLaVA-1.5 \\ + SBERT + Flan-T5} & 66.08 & 72.60 & 72.60 \\

\hline
\end{tabular}
\end{table}
{Table \ref{tab3} summarizes the performance of the best-performing large model-based solution for each adaptation method. The results demonstrate that large models exhibit strong capabilities in addressing the task of social media image categorization into CES classes.
The highest performance was achieved using LVMs within the supervised learning setting, followed by LVLMs under the in-context learning paradigm. The top-performing model in our study was DINOv2, an LVM utilizing the ViT-L/14 backbone, which achieved an accuracy of 97.08\%. This was followed by GPT-4o, an LVLM, which reached an accuracy of 88.85\%. Among all adaptation strategies, lightweight fine-tuning proved to be the most effective.}

{These findings suggest that while LVMs can deliver the highest accuracy, they require more extensive customization through fine-tuning. In contrast, LVLMs offer a more lightweight and user-friendly alternative—requiring no programming or AI expertise—by relying solely on prompt engineering. While this approach yields promising results, it comes with a  8.23\% decrease in accuracy compared to the best-performing LVM.}

{The class-specific results in Table \ref{tab4} indicate that the LVLM GPT-4o, adapted via prompting, achieved the highest performance in three out of six CES classes: Fauna-Flora, Gastronomy, and Landscape-Nature, with accuracies of 99.38\%, 98.75\%, and 98.75\%, respectively. In contrast, the LVM DINOv2, adapted using lightweight fine-tuning, performed best in identifying the Cultural-Religious and Urban-Rural classes, both with an accuracy of 97.50\%. Lastly, the Sports class was most accurately recognized by the hybrid solution combining LVLMs and LLMs through unsupervised clustering and zero-shot learning, also achieving an accuracy of 97.50\%.}

These findings reveal that each category achieved optimal classification performance with a different model, highlighting the potential of leveraging an ensemble approach to improve overall classification accuracy. An ensemble could combine the strengths of various models, allowing for enhanced performance across diverse categories. However, this strategy would require significantly higher computational resources, making it both costly and environmentally taxing due to the increased CO2 emissions associated with extended processing time. It is crucial to consider this trade-off to optimize both accuracy and sustainability while using large uni- and multi-modal models.

\begin{table}[H]
\caption{{Class-specific test results of the best-performing large model-based solution for each adaptation method. The top-1 results are highlighted in bold, and the top-2 results are underlined.}}
\label{tab4}
\resizebox{\textwidth}{!}{%
\begin{tabular}{llrrrrrr}
\hline
 \multicolumn{1}{c}{\textbf{\thead{Adaptation method}}} & \multicolumn{1}{c}{\textbf{\thead{Model}}} & \multicolumn{1}{c}{\textbf{\thead{Cultural\\Religious}}} & \multicolumn{1}{c}{\textbf{\thead{Fauna\\Flora}}} & \multicolumn{1}{c}{\textbf{\thead{Gastronomy}}} & \multicolumn{1}{c}{\textbf{\thead{Landscape\\Nature}}} & \multicolumn{1}{c}{\textbf{\thead{Sports}}} & \multicolumn{1}{c}{\textbf{\thead{Urban\\Rural}}} \\ \hline

\thead{Lightweight Fine-Tuning} & \thead{DINOv2 - ViT-L/14} & \bf 97.50 & \underline{98.12} & \underline{98.12} & \underline{98.12} & \underline{93.12} & \bf 97.50 \\

\thead{Zero-Shot Prompting} & \thead{GPT-4o (No Batch)} & 78.75 & \bf 99.38 & \bf 98.75 & \bf 98.75 & 63.75 & 93.75 \\ 

\thead{Few-Shot} & \thead{DINOv2 - ViT-S/14} & 79.79 & 76.80 & 94.06 & 83.13 & 77.25 & 90.23 \\

\thead{Dimension Reduction + \\Clustering + Zero-Shot} & \thead{KMeans + LLaVA-1.5 \\ + SBERT + Flan-T5} & 0 & 66.88 & 98.12 & 93.65 & \bf 97.50 & 79.38 \\ \hline

\hline
\end{tabular}%
}
\end{table}

The costs associated with analyzing our dataset using the different variants and settings of GPT-4 (used through ChatGPT) are outlined in Table \ref{tab_prices}. As shown, the pricing depends on several factors: the GPT-4 variant used, the execution mode, and the number of input and output tokens processed. Although  OpenAI documentation highlights that GPT-4o-min is  cheaper than the larger GPT-4o, our analysis shows that for image processing and with the same prompt GPT-4o-min automatically generates $20\times$ more input tokens than GPT-4o and hence using GPT-4o-mini ends to be more expensive as the final price increases linearly with the total number of tokens. 
 
\begin{table}[H]
\centering
\caption{ The prices of analyzing our dataset with GPT-4 (used through ChatGPT) in terms of the used model, execution mode and number of used input and output tokens. All other models are freely available.} 
\label{tab_prices}
\begin{tabular}{lcllrrrr} \hline
\bf Model  & \bf Prompt	& \bf Mode	& \bf updated	& \textbf{\thead{Input \\tokens}} &	\textbf{\thead{Output \\tokens}} &	\textbf{\thead{Input cost\\ (\$USD)}} &	\textbf{\thead{Output cost \\(\$USD)}} \\\hline
\multirow{4}{*}{GPT-4o-mini} & \multirow{2}{*}{1}	& Batch  & 2024-Jul-18	&20195532&	15537&	 1.5146&	0.0046	 \\
&	& No Batch  & 2024-Jul-18	&20195532&	15537&	 3.0293 & 0.0093	\\
& \multirow{2}{*}{2}	& Batch & 2024-Jul-18	&20335692&	26132&	1.5251	&0.0078\\
&	& No Batch	& 2024-Jul-18	&20335692&	26132&	 3.0503	&0.0156\\ \hline
\multirow{4}{*}{GPT-4o} & \multirow{2}{*}{1} & Batch	& 2024-Aug-06&	640860	&13331	 &	0.8010&0.0666	\\
&	& No Batch	& 2024-Aug-06&	640860	&13331	 &	1.6021	&0.1333\\
& \multirow{2}{*}{2}	& Batch	& 2024-Aug-06&	788840	&12116	 &	0.9860	&0.0605\\
&	& No Batch	& 2024-Aug-06&	788840	&12116	 &	1.9721	&0.1211\\

\hline
\end{tabular}

\end{table}

\section{Conclusion} \label{S6}
{This study addressed the challenge of categorizing social media images into a set of Cultural Ecosystem Services (CES) classes by leveraging recent advances in large models, including LVLMs, LVMs, and LLMs. A range of adaptation strategies was explored, spanning from supervised learning via lightweight fine-tuning, to in-context learning using prompting and few-shot learning, as well as a hybrid approach combining unsupervised and in-context setting using dimensionality reduction followed by clustering paired with zero-shot learning.}  

{The results confirm that large models are highly capable of solving the CES classification task. The best overall performance was achieved using the LVM DINOv2, adapted through lightweight fine-tuning, followed closely by the LVLM GPT-4o, which performed well under a zero-shot prompting setting.
Class-specific analysis revealed that optimal accuracy across all CES classes could potentially be achieved through an ensemble of large models. However, this approach comes with increased computational costs and environmental impact due to higher CO2 emissions.}

{Overall, while LVMs offer superior accuracy, they require more complex customization and fine-tuning. In contrast, LVLMs provide a lightweight and user-friendly alternative—requiring no programming or AI background—by relying on prompt engineering, achieving competitive results with only an 8.23\% reduction in accuracy compared to the best-performing LVM.}

{For future work, and inspired by the success of domain-specific CNN-based models in fields such as medical imaging \cite{esteva2017dermatologist}, we plan to explore CES classification using finer-grained categories. We also aim to incorporate multimodal social media data—combining both textual and visual inputs—to enrich model understanding and performance.}


\section*{Declaration of Competing Interest}
The authors declare that they have no known competing financial interests or personal relationships that could have appeared to influence the work reported in this paper.

\section*{Data availability}
This dataset (Version 1.0) \citep{khaldi_2024_14035313} is available to the public through an unrestricted data repository hosted by \href{https://zenodo.org/}{Zenodo} at: \blue{\url{https://zenodo.org/records/14035313}}


\section*{Acknowledgements}
This work was supported by EarthCul project (reference PID2020-118041GB-I00), funded by the Spanish Ministry of Science and Innovation. It was also part of the project BIOD22\_002 funded by  la Consejería de Universidad, Investigación e Innovación y el Gobierno de España y por la Unión Europea - NextGenerationEU. We sincerely acknowledge Ricardo Moreno-Llorca for his valuable help in dataset collection, and A. Sofia Vaz and A. Sofia Cardoso for their encouragement in the use of artificial intelligence on social media data for cultural ecosystem services assessment. JML was funded by the Plan Propio de Investigación (P9) of the University of Granada. This study has received credits from the OpenAI Researcher Access Program. This article contributes to the GEOBON research on ecosystem services.
\bibliography{sample}

\end{document}